\documentclass{article}

\usepackage{microtype}
\usepackage{graphicx}
\usepackage{booktabs}
\usepackage{hyperref}

\usepackage[accepted]{icml2026}

\usepackage{amsmath}
\usepackage{amssymb}

\icmltitlerunning{Majority Vote Silences Minority Values}

\begin{document}

\twocolumn[
\icmltitle{Majority Vote Silences Minority Values: Annotator Disagreement\\
           at the Hate/Offensive Boundary in HateXplain}

\begin{icmlauthorlist}
\icmlauthor{Joshua Muhumuza}{mak}
\icmlauthor{Joab Agaba Ezra}{mak}
\icmlauthor{Mercy Amiyo}{mak}
\end{icmlauthorlist}
\icmlaffiliation{mak}{Makerere University, Kampala, Uganda}
\icmlcorrespondingauthor{Joshua Muhumuza}{joshua.muhumuza@mak.ac.ug}

\icmlkeywords{pluralistic alignment, annotator disagreement, hate speech,
              soft labels, majority vote, value pluralism, BERT}

\vskip 0.3in
]

\printAffiliationsAndNotice{}

%%%%%%%%%%%%%%%%%%%%%%%%%%%%%%%%%%%%%%%%%%%%%%%%%%%%%%%%%%%%%%%%%%%
\begin{abstract}
Hate speech annotation pipelines routinely collapse annotator disagreement
into majority vote labels before training. We show that this aggregation
is not neutral: 42.6\% of all annotator disagreement in HateXplain
concentrates specifically at the hate/offensive boundary, a pattern
consistent with annotators applying different thresholds for where hate
begins ($\chi^{2} = 135.199$, $df = 2$, $p < 0.0001$). Both a
hard-label BERT model~(Model~A) and a soft-label model~(Model~B) drop
22~percentage points in accuracy from agreed posts ($\sim$80\%) to
disagreement posts ($\sim$58\%), confirmed at $p < 0.0001$. A
per-annotator multi-head model~(Model~C) widens this gap further to
28~points while collapsing offensive disagreement accuracy to 0.245.
Critically, Model~A expresses significantly higher confidence on boundary
case errors than Model~C (0.710~vs.\ 0.495, $p < 0.0001$), meaning
standard evaluation metrics will not detect the failure. Three downstream
interventions of increasing sophistication all fail to recover boundary
accuracy. We argue the problem is structural. Majority vote presents a
contested judgment as ground truth, and models inherit that
false certainty. The intervention must be upstream in annotation design.
\end{abstract}

%%%%%%%%%%%%%%%%%%%%%%%%%%%%%%%%%%%%%%%%%%%%%%%%%%%%%%%%%%%%%%%%%%%
\section{Introduction}
\label{sec:intro}

Hate speech detection isn't a purely empirical or technical category like
"detecting spam" or "identifying cats in images.". Whether a post is hateful or
merely offensive is not an objective fact. It is a judgment that depends on
who is reading, what community they belong to, and what threshold they hold
for harm. Yet standard annotation pipelines treat this judgment as if it were
objective: crowdworkers label each post, the majority wins, and the resulting
label is presented to a model as ground truth. \citet{plank2022problem} terms
this assumption the ``problem'' of human label variation: treating genuine
variation as noise to be minimised rather than signal to be preserved.

This paper asks what is lost when that majority wins. Rather than propose a
new model or training objective, we locate precisely where in HateXplain
annotator disagreement concentrates, characterise what that concentration
means for model behaviour, and show that three downstream interventions of
increasing sophistication all fail to recover what majority vote discards.

Our central finding is that 42.6\% of all annotator disagreement in HateXplain
occurs specifically at the hate/offensive boundary ($\chi^{2} = 135.199$,
$df = 2$, $p < 0.0001$). The concentration is not random noise but a
structural pattern at the boundary where annotators must decide whether
content crosses from offensiveness into hatred. That decision depends on
where each annotator places the threshold between offence and hatred,
and majority vote resolves it by silencing whichever annotators land on
the minority side.

The consequence for deployed models is concerning. Both
Models~A and B drop over 22~percentage points in accuracy from agreed to
disagreement posts ($p < 0.0001$). Model~C (per-annotator heads, the
strongest baseline we test) widens the gap to 28~points and collapses
offensive disagreement accuracy to 0.245. Model~A also expresses
significantly higher confidence on boundary case errors than
Model~C (0.710~vs.\ 0.495, $p < 0.0001$). A system that is wrong
\emph{and certain} on culturally contested inputs is a dangerous
failure mode that standard accuracy metrics will not detect.

\paragraph{Contributions.}
\textbf{(1)}~Annotator disagreement in HateXplain is not uniformly
distributed: 42.6\% concentrates at the hate/offensive value boundary
($\chi^{2}=135.199$, $df=2$, $p<0.0001$), with offensive posts
disagreeing at 67.9\% versus 35.5\% for normal posts.
\textbf{(2)}~A 22-point accuracy gap between agreed and disagreement posts
persists across all three training regimes ($p<0.0001$).
\textbf{(3)}~Three downstream interventions (hard labels, soft labels, and
per-annotator heads) all fail to recover boundary accuracy; the gap widens
under the strongest baseline.
\textbf{(4)}~Hard-label models are significantly overconfident on boundary
case errors relative to per-annotator models (0.710~vs.\ 0.495,
$p<0.0001$), confirming the problem is structural, not architectural.

%%%%%%%%%%%%%%%%%%%%%%%%%%%%%%%%%%%%%%%%%%%%%%%%%%%%%%%%%%%%%%%%%%%
\section{Related Work}
\label{sec:related}

\paragraph{Disagreement in annotation.}
\citet{uma2021learning} survey learning from disagreement across NLP tasks,
showing that soft label approaches generally improve calibration.
\citet{plank2022problem} reframes annotator disagreement as human label
variation (genuine signal rather than noise) and catalogues approaches
that preserve it. \citet{davani2022dealing} make the closest argument to
ours, showing that majority vote loses signal on subjective tasks and that
per-annotator models recover minority perspectives.
\citet{gordon2021disagreement} introduce the disagreement deconvolution,
showing that standard ML metrics systematically overstate performance on
subjective tasks because they evaluate against aggregated labels rather than
individual annotators, a finding that motivates our subgroup analysis.
\citet{gordon2022jury} extend this with jury learning, an architecture
that models every annotator and lets practitioners explicitly weight
demographic groups at inference. Our work differs in three ways: we
\emph{measure} where disagreement concentrates rather than proposing a new
aggregation mechanism; we show the concentration is category-specific with
precise statistics; and we test the per-annotator approach directly,
finding it widens rather than closes the agreement gap on boundary cases.

\paragraph{Hate speech detection.}
\citet{mathew2021hatexplain} introduced HateXplain, covering hatespeech,
offensive speech, and normal content with rationale annotations. They note
the hate/offensive distinction is the hardest classification boundary but
do not quantify what proportion of disagreement it accounts for, nor its
consequences for model confidence. \citet{leonardelli2021agreeing} study
disagreement in sentiment annotation and argue for preserving annotator
perspectives, but do not examine overconfidence as a downstream consequence.
\citet{fleisig2023majority} argue that when the demographic group targeted
by a statement is a minority of the annotator pool, the majority vote may
be systematically wrong, and show that modelling target-group ratings
recovers signal that aggregation discards. Our work connects these threads
empirically.

\paragraph{Pluralistic alignment.}
\citet{sorensen2024roadmap} define pluralistic alignment as the challenge
of building AI systems that represent diverse human values rather than a
majority consensus. \citet{aroyo2023dices} operationalise this concern for
conversational safety with the DICES dataset, which preserves rater
demographics and uses a graduated safety scale to enable explicit study of
how aggregation choices affect outcomes. Current RLHF pipelines face an
analogous aggregation problem: human preference comparisons are collapsed
into reward signals that may encode majority rater values as universal.
We do not test RLHF systems here, but our findings provide concrete
grounding for this theoretical concern and suggest testable hypotheses
for future work.

%%%%%%%%%%%%%%%%%%%%%%%%%%%%%%%%%%%%%%%%%%%%%%%%%%%%%%%%%%%%%%%%%%%
\section{Methods}
\label{sec:methods}

\subsection{Dataset}

We use HateXplain \citep{mathew2021hatexplain}, loaded from the original
GitHub repository to preserve raw annotator labels. Each of 20,148~posts
is annotated by three crowdworkers with one of three labels:
\texttt{hatespeech}, \texttt{normal}, or \texttt{offensive}. We use the
official train/validation/test splits. All classification experiments are
conducted on the test set ($n=1{,}892$); disagreement characterisation
statistics are computed over the full dataset.

We define \emph{disagreement} as any post where annotators did not
unanimously agree (unique label count~$>1$). \emph{Boundary cases} are
disagreement posts where raw labels are a subset of
\{hatespeech, offensive\}: posts where no annotator chose normal, but
annotators disagreed between the two harm categories. Soft label
distributions are computed directly from raw annotator labels before any
aggregation.

\subsection{Models}

We train three BERT-base-uncased \citep{devlin2019bert} classifiers of
increasing sophistication.

\textbf{Model~A (Hard)} is trained on majority vote labels using standard
cross-entropy loss, the default pipeline for hate speech classification.

\textbf{Model~B (Soft)} is trained on soft labels derived from the annotator
distribution (e.g.\ a 2-1 hatespeech/offensive split yields
$[0.67,\,0.0,\,0.33]$) using KL divergence loss.

\textbf{Model~C (Per-Annotator)} uses a shared BERT encoder with one
separate classification head per qualifying annotator (106~annotators with
$\geq$50 training posts each). During training, each sample is routed to
its annotator's specific head. At inference, all heads are ensembled by
averaging their softmax probabilities. This follows the approach of
\citet{davani2022dealing} and constitutes our strongest downstream baseline.

All models are trained for 3~epochs with AdamW ($lr=2{\times}10^{-5}$),
linear warmup over 10\% of steps, and gradient clipping at 1.0.

\subsection{Statistical Analysis}

Group comparisons use the Mann-Whitney U~test (two-tailed, $\alpha=0.05$).
Chi-square tests assess whether disagreement rates differ across label
categories. We report exact test statistics, degrees of freedom, and
$p$-values for all claims.

%%%%%%%%%%%%%%%%%%%%%%%%%%%%%%%%%%%%%%%%%%%%%%%%%%%%%%%%%%%%%%%%%%%
\section{Results}
\label{sec:results}

\subsection{Disagreement Concentrates at the Value Boundary}

Of 15,383~training posts, 48.7\% show annotator disagreement, making it
a structural feature of the dataset rather than a marginal one. Among
test set disagreement posts ($n=950$), 42.6\% are boundary cases where
annotators chose exclusively between hatespeech and offensive labels
($n=405$).

\begin{figure}[h]
\vskip 0.2in
\begin{center}
\centerline{\includegraphics[width=\columnwidth]{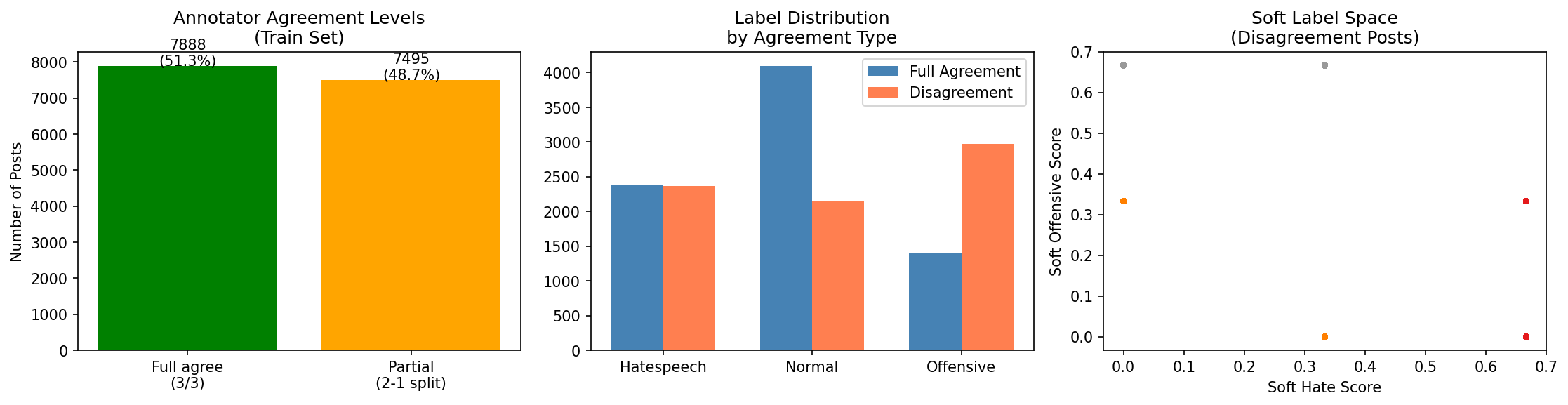}}
\caption{Left: annotator agreement levels in the training set (48.7\%
show disagreement). Centre: label distribution by agreement type.
Offensive posts are over-represented in disagreement cases. Right:
soft label space for disagreement posts, showing discrete probability
mass concentrated at the hate/offensive boundary.}
\label{fig:disagreement}
\end{center}
\vskip -0.2in
\end{figure}

A chi-square test confirms that disagreement rate differs significantly
across label categories ($\chi^{2}=135.199$, $df=2$, $p<0.0001$). Offensive
posts disagree at 67.9\% (372/548), hatespeech posts at 50.5\% (300/594),
and normal posts at 35.5\% (278/782). The hate/offensive boundary is where
annotator judgments most systematically diverge, not merely the hardest
classification problem. This concentration is consistent with genuine
threshold disagreement among annotators: disagreement clusters at the
exact point where distinguishing harm categories requires a judgment call
about severity (Figure~\ref{fig:disagreement}).

\subsection{The Agreement Gap Persists Across All Three Models}

\begin{figure}[h]
\vskip 0.2in
\begin{center}
\centerline{\includegraphics[width=\columnwidth]{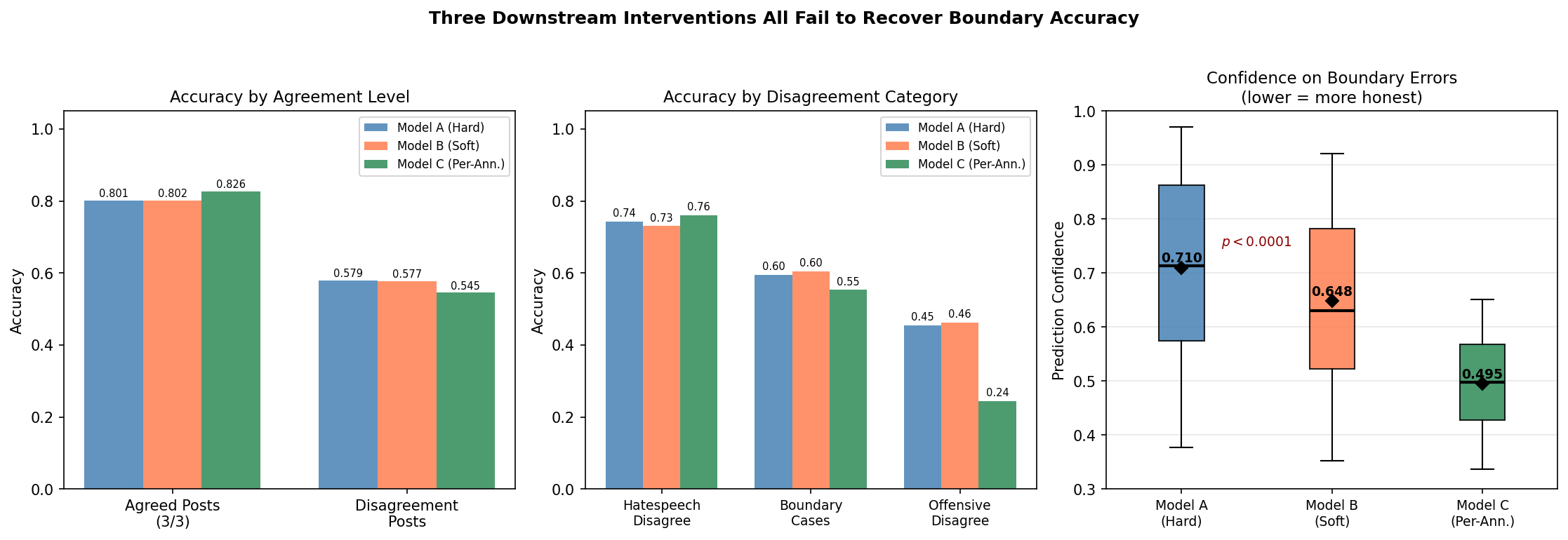}}
\caption{Three downstream interventions all fail to recover boundary
accuracy. \textbf{Left}: accuracy by agreement level. Agreed post
accuracy is stable across models ($\sim$80\%) while disagreement post
accuracy falls further under Model~C (0.545 vs.\ 0.579 for Model~A),
widening the gap to 28 points. \textbf{Centre}: accuracy by disagreement
category. Hatespeech disagreement is relatively stable across models,
but offensive disagreement collapses to 0.245 under per-annotator heads,
confirming the offensive boundary is the critical failure point.
\textbf{Right}: confidence on boundary errors (box plots, mean marked
with diamond). Model~A is significantly overconfident (0.710) while
Model~C approaches appropriate uncertainty (0.495, $p<0.0001$), but
calibration gains come at the cost of accuracy.}
\label{fig:three_models}
\end{center}
\vskip -0.2in
\end{figure}

Table~\ref{tab:results} and Figure~\ref{fig:three_models} present the core
results. All three models perform substantially better on agreed posts
($\sim$80\%) than on disagreement posts ($\sim$55--58\%). The agreement
gap is confirmed at $p<0.0001$ for all three models by Mann-Whitney U~test.
Standard overall accuracy (69--70\%) completely obscures this gap, echoing
the gap-hiding pattern \citet{gordon2021disagreement} document for
aggregated metrics on subjective tasks.

The gap \emph{widens} under the strongest baseline. Model~C
(per-annotator heads) achieves the best agreed-post accuracy (0.826) but
the worst disagreement-post accuracy (0.545), producing the largest
agreement gap of 28.1~points. Offensive disagreement accuracy collapses
to 0.245 under Model~C, less than half of Model~A's already-poor 0.454.
No downstream intervention recovers the boundary.

\begin{table}[h]
\caption{Results across all three models on HateXplain test set
  ($n=1{,}892$). n.s.\ = not significant at $\alpha=0.05$.
  Minority alignment discussed in Section~\ref{sec:perann}.}
\label{tab:results}
\vskip 0.1in
\begin{center}
\begin{small}
\begin{tabular}{lcccc}
\toprule
Metric & Model A & Model B & Model C & Sig. \\
       & (Hard)  & (Soft)  & (Per-Ann.) & \\
\midrule
Overall accuracy    & 0.691 & 0.691 & 0.688 & --   \\
Agreed posts        & 0.801 & 0.802 & 0.826 & --   \\
Disagreement posts  & 0.579 & 0.577 & 0.545 & $p{<}0.0001$ \\
Agreement gap       & 0.222 & 0.225 & 0.281 & --   \\
\midrule
Boundary cases      & 0.595 & 0.605 & 0.553 & n.s. \\
Hatespeech disagree & 0.743 & 0.730 & 0.760 & n.s. \\
Offensive disagree  & 0.454 & 0.462 & 0.245 & --   \\
\midrule
Minority alignment  & 0.291 & 0.308 & 0.307 & \S\ref{sec:perann} \\
Conf.\ (bdry err.) & 0.710 & 0.648 & 0.495 & $p{<}0.0001$ \\
\bottomrule
\end{tabular}
\end{small}
\end{center}
\vskip -0.1in
\end{table}

\subsection{Soft Labels Do Not Resolve the Gap}
\label{sec:softlabels}

Model~B does not significantly outperform Model~A on any disagreement
subgroup. The boundary case difference (0.605~vs.\ 0.595) yields $p=0.284$.
Hatespeech disagreement (0.730~vs.\ 0.743) yields $p=0.236$. Offensive
disagreement marginally favours Model~B (0.462~vs.\ 0.454) but $p=0.152$.
None survive significance testing.

\subsection{Per-Annotator Heads: More Honest, Not More Accurate}
\label{sec:perann}

Model~C produces the most informative pattern of the three. On agreed
posts it achieves the best accuracy (0.826), confirming the shared encoder
learns strong general representations. But on disagreement posts it performs
worst (0.545), and on offensive disagreement it collapses to 0.245. The
per-annotator heads have learned individual annotator decision boundaries
that do not generalise to the contested cases where annotators genuinely
disagree.

The one meaningful gain from Model~C is calibration. Boundary error
confidence drops progressively from Model~A (0.710) to Model~B (0.648) to
Model~C (0.495), confirmed at $p<0.0001$ (Figure~\ref{fig:three_models},
right). Model~C is appropriately uncertain on the posts it cannot classify,
though that uncertainty comes at the cost of accuracy elsewhere.

\begin{figure}[h]
\vskip 0.2in
\begin{center}
\centerline{\includegraphics[width=\columnwidth]{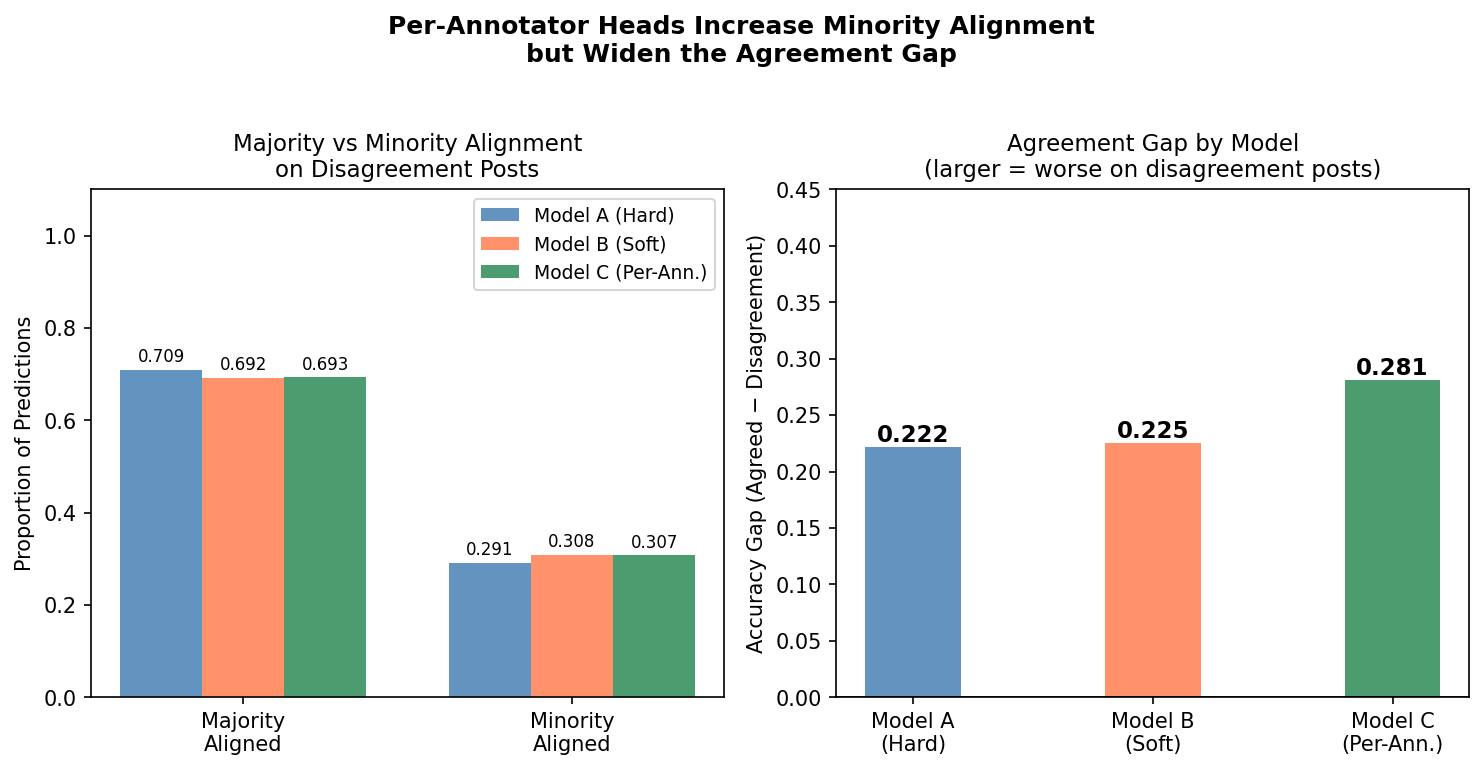}}
\caption{Left: majority vs.\ minority alignment rates on disagreement
posts. All three models remain strongly majority-aligned ($\sim$70\%).
Right: agreement gap by model. The gap widens monotonically from
Model~A to Model~C, confirming no downstream intervention closes it.}
\label{fig:minority}
\end{center}
\vskip -0.2in
\end{figure}

The minority alignment rate tells a similarly nuanced story
(Figure~\ref{fig:minority}). All three models are strongly majority-aligned:
roughly 70\% of predictions on disagreement posts match the majority label
and only 30\% align with any minority annotator. The agreement gap widens
monotonically from Model~A (0.222) to Model~B (0.225) to Model~C (0.281),
confirming that no downstream intervention closes the structural gap that
majority vote creates.

\paragraph{A note on the inference rule.}
Model~C ensembles its per-annotator heads by averaging softmax
probabilities, the inference rule used in the original per-annotator
proposal we follow \citep{davani2022dealing}. Mean-pooling has a known
limitation in our setting: it re-aggregates the per-annotator signal at
inference time, which can wash out minority predictions even when
individual heads have learned them. A natural stronger pluralistic
baseline is jury learning \citep{gordon2022jury}, in which practitioners
can explicitly weight annotator demographic groups at inference rather
than averaging uniformly. We do not test jury learning here, both because
HateXplain lacks the annotator demographic metadata jury learning requires
and because our claim is about the failure of \emph{standard} downstream
interventions. We therefore frame Model~C as a strong baseline for the
default per-annotator approach, not the strongest possible pluralistic
baseline. Implementing a jury-learning-style readout on a demographically
labelled dataset is the natural next test (Section~\ref{sec:limitations}).

\subsection{Boundary Disagreement Is Not Driven by Annotation Error}
\label{sec:rationale}

A key alternative explanation for our boundary concentration finding is
annotation error: annotators misreading posts rather than genuinely
disagreeing about their meaning. HateXplain's token-level rationale
annotations let us test the prediction this explanation makes. For each
boundary disagreement post, we compute the Jaccard similarity between the
token sets highlighted by pairs of annotators as justification for their
label choice.

\begin{figure}[h]
\vskip 0.2in
\begin{center}
\centerline{\includegraphics[width=\columnwidth]{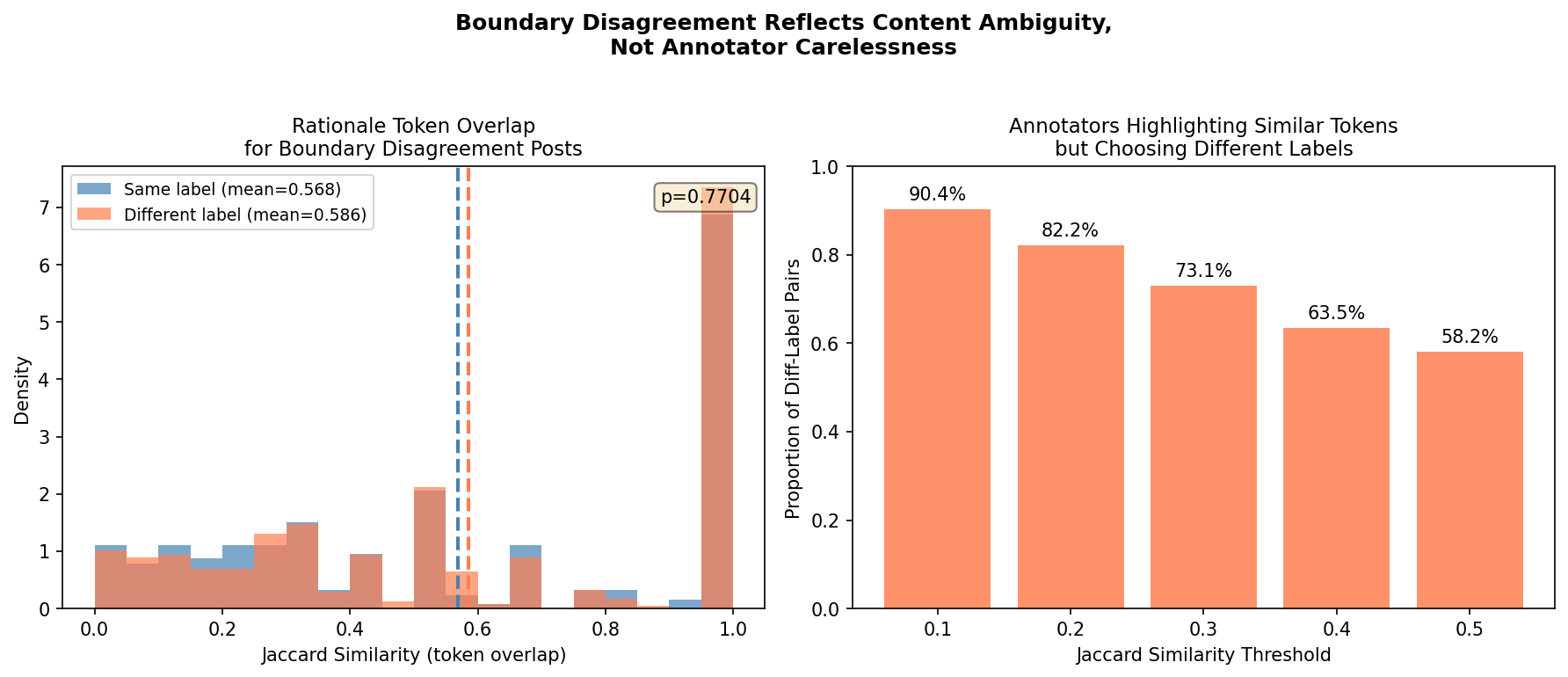}}
\caption{\textbf{Left}: distribution of token overlap (Jaccard similarity)
between annotator pairs on boundary disagreement posts. Annotators who
chose \emph{different} labels (coral) show equal or higher token overlap
than those who chose the \emph{same} label (blue, $p=0.770$); the
difference is not significant, but the direction is inconsistent with an
annotation-error account. \textbf{Right}: 73.1\% of different-label
pairs highlight substantially overlapping tokens (Jaccard~$\geq 0.3$),
and 58.2\% overlap at Jaccard~$\geq 0.5$. A clear mode at
Jaccard~$=1.0$ in the left panel captures pairs that highlighted
identical tokens but reached different labels, the cleanest case of
threshold disagreement on shared evidence.}
\label{fig:rationale}
\end{center}
\vskip -0.2in
\end{figure}

The strongest evidence against the annotation error account is that
the majority of different-label pairs highlight substantially overlapping
tokens. Of all different-label pairs in our boundary set, 73.1\% have
Jaccard similarity $\geq 0.3$ and 58.2\% have Jaccard $\geq 0.5$
(Figure~\ref{fig:rationale}, right). A visible mode at exactly
Jaccard~$=1.0$ (Figure~\ref{fig:rationale}, left) corresponds to pairs of
annotators who highlighted \emph{identical} token sets yet assigned
different labels: the cleanest possible signature of threshold
disagreement, since the same evidence is being weighed and reaching
different conclusions. We do not claim these cases prove the absence of
all noise; we claim that the noise model under which annotators are
inattentively misreading the post is inconsistent with annotators
highlighting the same content.

This is corroborated, less strongly, by the full distribution comparison.
Annotators who chose \emph{different} labels show mean token overlap of
0.586, marginally \emph{higher} than annotators who chose the \emph{same}
label (0.568, Mann-Whitney $p=0.770$). The difference is not statistically
significant, so we do not treat it as evidence of value differences. We
note only that the direction is opposite to what an annotation-error
account would predict (which would expect lower overlap among
different-label pairs). Combined with the perfect-overlap mode, the
overall picture is one in which annotators read the same content and
reach different conclusions about how to label it, a pattern more easily
explained by differing thresholds than by carelessness. We cannot, with
three annotators per post and no demographic metadata, distinguish
threshold differences rooted in cultural background from threshold
differences rooted in idiosyncratic interpretation; we return to this
limitation in Section~\ref{sec:limitations}.

%%%%%%%%%%%%%%%%%%%%%%%%%%%%%%%%%%%%%%%%%%%%%%%%%%%%%%%%%%%%%%%%%%%
\section{Discussion}
\label{sec:discussion}

\subsection{Three Baselines, One Conclusion}

Our three-model comparison produces a convergent result. Across
majority vote labels, soft label distributions, and per-annotator head
ensembles, no downstream intervention recovers accuracy on hate/offensive
boundary cases. The agreement gap does not shrink; it grows. The only
dimension that improves is calibration: Model~C is appropriately uncertain
on boundary inputs. But a model that is uncertain and inaccurate is not
a deployed solution, only evidence that the task as formulated cannot be
solved at the training objective level.

This convergent failure across three baselines of increasing sophistication
constitutes stronger evidence for upstream intervention than any single
baseline failure would. The problem is not a weak choice of model; it is
that the label was never reliable on these inputs.

\subsection{What Upstream Intervention Means}

Three concrete upstream changes follow from our results. First, and most
directly testable, annotation schemas should replace the discrete
hate/offensive distinction with a graduated harm scale (e.g.\ 1--5).
This is falsifiable: a dataset annotated with a continuous scale should
show lower inter-annotator disagreement at the boundary and produce models
with smaller agreement gaps than the 22--28 points we observe here.
\citet{aroyo2023dices} provide preliminary evidence that graduated scales
combined with demographically diverse rater pools surface aggregation
effects that binary schemas hide. Second, high-disagreement posts should
be flagged at inference time: Model~C's lower confidence (0.495) shows
this signal exists in the model; surfacing it to downstream users would
allow human review of contested inputs rather than silent automated
decisions. Third, annotation pools should represent the target communities
whose content is moderated: content about LGBTQ, religious minority, and
racial minority communities (all present in HateXplain) should be reviewed
by annotators from those communities, whose value judgments about where
offence ends and hatred begins may systematically differ from majority
annotator pools. \citet{fleisig2023majority} provide direct empirical
support: when target-group members are a minority of raters, modelling
their ratings explicitly recovers signal that majority vote discards.

This third recommendation comes with real ethical tradeoffs that we do
not want to elide. Asking members of marginalised communities to annotate
hateful content directed at their own communities exposes them to
material harm: repeated exposure to slurs and threats is psychologically
costly, particularly when uncompensated or under-compensated. Operational
questions about who legitimately represents a community, how to balance
lived experience against exposure risk, and how to obtain meaningful
informed consent are not solved by saying ``recruit target-community
annotators.'' We take the recommendation to require, at minimum,
appropriate compensation, opt-in participation with clear content
warnings, mental-health support, and care around essentialising who
speaks for a community. The aim of pluralistic annotation is to include
affected voices on their own terms, not to redistribute the burden of
moderation work onto the people most harmed by the underlying content.

\subsection{Implications for Pluralistic AI: A Hypothesis}

We hypothesise that the aggregation problem we document is not
domain-specific. RLHF reward models collapse human preference comparisons
into scalar signals; safety classifiers and LLM safety benchmarks
collapse harm judgments into hard labels. In each case, averaging
presents a value-laden judgment as if it were a fact, and our results
suggest this produces overconfident models on exactly the inputs where
confidence is least warranted. We stress that the evidence we present
covers one boundary in one dataset under one model family; this is a
hypothesis to be tested, not a demonstrated property of value-laden
aggregation in general. Testing it on graduated-scale safety datasets
such as DICES \citep{aroyo2023dices}, on additional model families, and
on RLHF reward models directly are all natural next steps.

\subsection{Limitations}
\label{sec:limitations}

With three annotators per post we cannot distinguish genuine threshold
differences (whether cultural or idiosyncratic) from annotation noise. A
2-1 split could reflect one annotator misreading the post. Future work
with annotator demographic metadata and larger annotator pools
\citep[e.g.][]{aroyo2023dices} could test whether boundary disagreements
correlate with annotator background, and could implement minority-aware
inference rules such as jury learning \citep{gordon2022jury} that
HateXplain's metadata does not currently support. All experiments use
HateXplain with BERT-base; generalisation to other datasets and model
families remains to be tested.

%%%%%%%%%%%%%%%%%%%%%%%%%%%%%%%%%%%%%%%%%%%%%%%%%%%%%%%%%%%%%%%%%%%
\section{Conclusion}
\label{sec:conclusion}

We identify a structural property of hate speech annotation with direct
consequences for pluralistic alignment: 42.6\% of annotator disagreement
in HateXplain concentrates at the hate/offensive boundary
($\chi^{2}=135.199$, $df=2$, $p<0.0001$), a pattern consistent with
annotators applying different thresholds for where hate begins. Three
BERT models of increasing sophistication (majority vote labels, soft
labels, and per-annotator head ensembles) all fail to recover accuracy
on boundary cases, and the agreement gap widens under the strongest
baseline (0.222 to 0.281). The only consistent improvement across models
is calibration: boundary error confidence drops from 0.710 (Model~A) to
0.495 (Model~C, $p<0.0001$), but appropriate uncertainty does not
produce correct predictions. Majority vote presents a contested judgment
as ground truth, models inherit that false certainty, and the
populations whose judgments were in the minority bear the cost. The
intervention must be upstream.

%%%%%%%%%%%%%%%%%%%%%%%%%%%%%%%%%%%%%%%%%%%%%%%%%%%%%%%%%%%%%%%%%%%
\section*{Impact Statement}
This paper examines how annotation aggregation methods affect hate speech
classifiers deployed in content moderation. Our findings have direct
implications for communities whose content is disproportionately moderated:
specifically, communities whose linguistic norms (including reclaimed
slurs, culturally specific expressions of offence, and in-group language)
are systematically misclassified at the hate/offensive boundary. HateXplain's
target community labels include African, LGBTQ, Muslim, Jewish, Hispanic,
and Asian communities, all of which are subject to moderation decisions made
by models trained on majority annotator judgments that may not represent
those communities' own norms \citep{fleisig2023majority}. By showing that
three downstream interventions fail to recover accuracy on contested inputs,
we argue that improvements require representation at the annotation stage,
not the model stage. No new models are deployed; all experiments use
existing public datasets and pretrained models. The datasets used contain
offensive and hateful content by design; researchers replicating this work
should be aware of this.

%%%%%%%%%%%%%%%%%%%%%%%%%%%%%%%%%%%%%%%%%%%%%%%%%%%%%%%%%%%%%%%%%%%
\bibliography{paper}
\bibliographystyle{icml2026}

\end{document}